\documentclass[10pt,twocolumn,letterpaper]{article}

\usepackage{cvpr}
\usepackage{times}
\usepackage{epsfig}
\usepackage{graphicx}
\usepackage{amsmath}
\usepackage{amssymb}
\usepackage{multirow}
\usepackage{makecell}
\usepackage{threeparttable}
\usepackage{booktabs}
\usepackage{stfloats}
\usepackage{url}
\usepackage{footnote}
\usepackage{subfig}

\usepackage{graphicx}
\usepackage{amsmath,amssymb} 
\usepackage[dvipsnames]{color} 
\usepackage{tabularx}

\usepackage[british,UKenglish,USenglish,english,american]{babel}

\usepackage[toc,page]{appendix}
\usepackage{comment}

\usepackage[bookmarks=false,colorlinks=true,linkcolor=black,citecolor=black,filecolor=black,urlcolor=black]{hyperref}

\newcommand{\ct}[1]{\fontsize{8pt}{1pt}\selectfont{#1}}
\newcolumntype{x}{>\small c}

\cvprfinalcopy 

\def\cvprPaperID{***} 

\begin{document}

\title{Convolutional Feature Masking for Joint Object and Stuff Segmentation}

\author{Jifeng Dai \qquad\qquad Kaiming He \qquad\qquad Jian Sun \vspace{8pt}\\
Microsoft Research\\
{\tt\small \{jifdai,kahe,jiansun\}@microsoft.com}
}

\maketitle

\begin{abstract}

The topic of semantic segmentation has witnessed considerable progress due to the powerful features learned by convolutional neural networks (CNNs) \cite{krizhevsky2012imagenet}. The current leading approaches for semantic segmentation exploit shape information by extracting CNN features from masked image regions. This strategy introduces artificial boundaries on the images and may impact the quality of the extracted features. Besides, the operations on the raw image domain require to compute thousands of networks on a single image, which is time-consuming.

In this paper, we propose to exploit shape information via masking convolutional features. The proposal segments (\eg, super-pixels) are treated as masks on the convolutional feature maps. The CNN features of segments are directly masked out from these maps and used to train classifiers for recognition. We further propose a joint method to handle objects and ``stuff'' (\eg, grass, sky, water) in the same framework. State-of-the-art results are demonstrated on benchmarks of PASCAL VOC and new PASCAL-CONTEXT, with a compelling computational speed.
\end{abstract}

\section{Introduction}

Semantic segmentation~\cite{kumar2005obj,JamieECCV06TextonBoost,yang2010layered,brox2011object} aims to label each image pixel to a semantic category. With the recent breakthroughs \cite{krizhevsky2012imagenet} by convolutional neural networks (CNNs) \cite{lecun1989backpropagation}, R-CNN based methods \cite{girshick2013rich,hariharan2014simultaneous} for semantic segmentation have substantially advanced the state of the art.

The R-CNN methods \cite{girshick2013rich,hariharan2014simultaneous} for semantic segmentation extract two types of CNN features - one is \emph{region features} \cite{girshick2013rich} extracted from proposal bounding boxes \cite{uijlings2013selective}; the other is \emph{segment features} extracted from the raw image content masked by the segments \cite{hariharan2014simultaneous}.
The concatenation of these features are used to train classifiers \cite{hariharan2014simultaneous}. These methods have demonstrated compelling results on this long-standing challenging task.

However, the raw-image-based R-CNN methods \cite{girshick2013rich,hariharan2014simultaneous} have two issues. First, the masks on the image content can lead to artificial boundaries. These boundaries do not exhibit on the samples during the network pre-training (\eg, in the 1000-category ImageNet \cite{deng2009imagenet}). This issue may degrade the quality of the extracted segment features. Second, similar to the R-CNN method for object detection \cite{girshick2013rich}, these methods need to apply the network on thousands of raw image regions with/without the masks. This is very time-consuming even on high-end GPUs.

The second issue also exists in R-CNN based object detection. Fortunately, this issue can be largely addressed by a recent method called SPP-Net \cite{he2014spatial}, which computes convolutional feature maps on the entire image only once and applies a spatial pyramid pooling (SPP) strategy to form cropped features for classification. The detection results via these cropped features have shown competitive detection accuracy \cite{he2014spatial}, and the speed can be $\sim$$50\times$ faster. Therefore, in this paper, we raise a question: \emph{for semantic segmentation, can we use the convolutional feature maps only?}

\begin{figure*}
  \centering
    \includegraphics[width=0.8\textwidth]{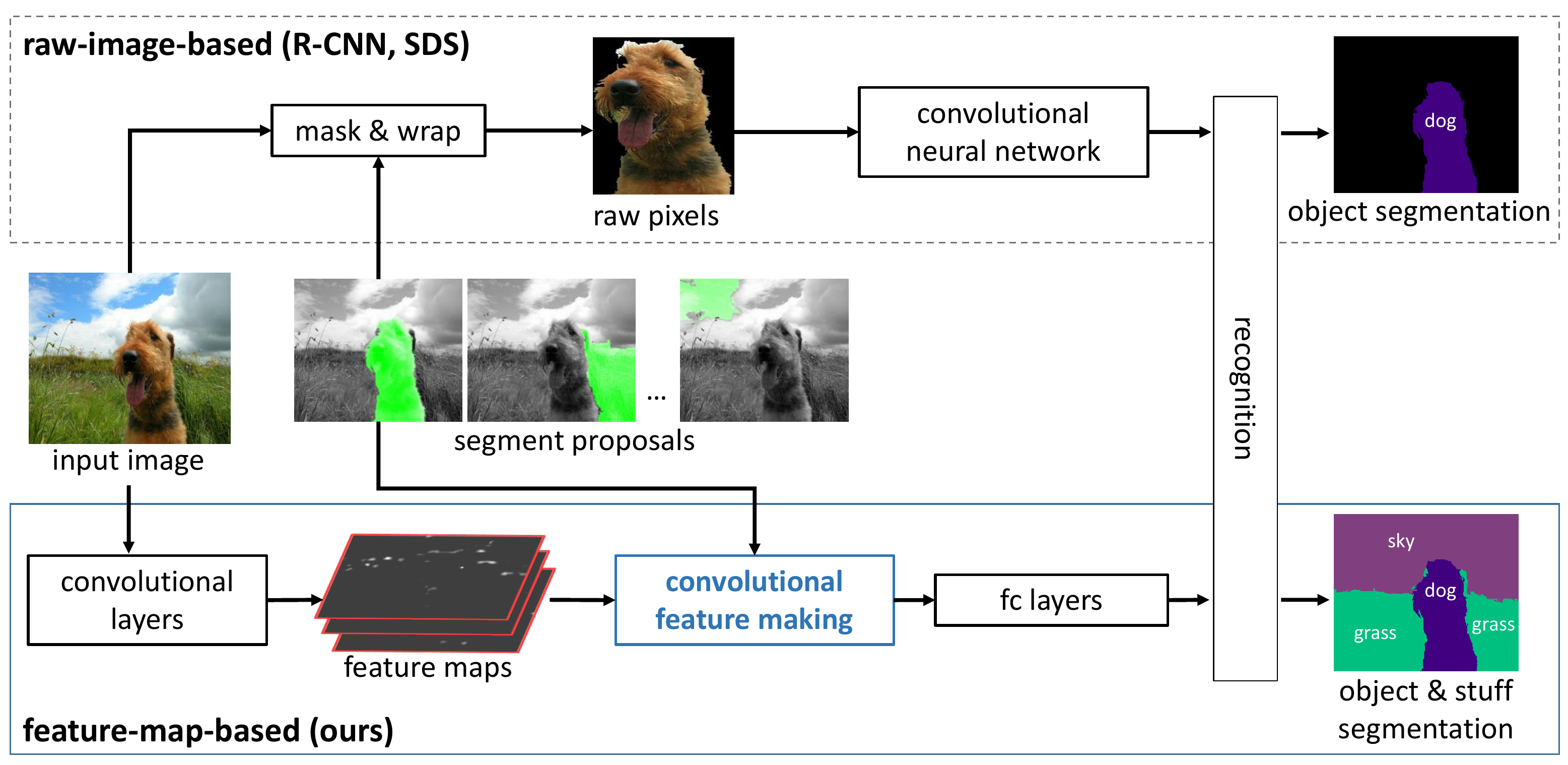}\\
\caption{System pipeline. \textbf{Top}: the methods of ``Regions with CNN features'' (R-CNN) \cite{girshick2013rich} and ``Simultaneous Detection and Segmentation'' (SDS) \cite{hariharan2014simultaneous} that operate on the raw image domain. \textbf{Bottom}: our method that masks the convolutional feature maps.}
\label{fig:algorithm_outline}
\end{figure*}

The first part of this work says \emph{yes} to this question. We design a \emph{convolutional feature masking} (CFM) method to extract \emph{segment features} directly from feature maps instead of raw images. With the segments given by the region proposal methods (\eg, selective search \cite{uijlings2013selective}), we project them to the domain of the last convolutional feature maps. The projected segments play as binary functions for masking the convolutional features. The masked features are then fed into the fully-connected layers for recognition. Because the convolutional features are computed from the unmasked image, their quality is not impacted. Besides, this method is efficient as the convolutional feature maps only need to be computed once. The aforementioned two issues involving semantic segmentation are thus both addressed. Figure~\ref{fig:algorithm_outline} compares the raw-image-based pipeline and our feature-map-based pipeline.

The second part of this paper further generalizes our method for joint object and stuff segmentation \cite{mottaghi2014role}. Different from objects, ``stuff'' \cite{mottaghi2014role} (\eg, sky, grass, water) is usually treated as the context in the image. Stuff mostly exhibits as colors or textures and has less well-defined shapes. It is thus inappropriate to use a single rectangular box or a single segment to represent stuff. Based on our masked convolutional features, we propose a training procedure that treats a stuff as a compact combination of multiple segment features. This allows us to address the object and stuff in the same framework.

Based on the above methods, we show state-of-the-art results on the PASCAL VOC 2012 benchmark \cite{everingham2010pascal} for object segmentation. Our method can process an image in a fraction of a second, which is $\sim$$150\times$ faster than the R-CNN-based SDS method \cite{hariharan2014simultaneous}.
Further, our method is also the \emph{first} deep-learning-based method ever applied to the newly labeled PASCAL-CONTEXT benchmark \cite{mottaghi2014role} for both object and stuff segmentation, where our result substantially outperforms previous states of the art.

\section{Convolutional Feature Masking}

\subsection{Convolutional Feature Masking Layer}

The power of CNNs as a generic feature extractor has been gradually revealed in the computer vision area \cite{krizhevsky2012imagenet,Donahue2013,zeiler2013visualizing,girshick2013rich,he2014spatial}. In Krizhevsky \etal's work \cite{krizhevsky2012imagenet}, they suggest that the features of the fully-connected layers can be used as \emph{holistic} image features, \eg, for image retrieval. In \cite{Donahue2013,zeiler2013visualizing}, these holistic features are used as generic features for full-image classification tasks in other datasets via transfer learning. In the breakthrough object detection paper of R-CNN \cite{girshick2013rich}, the CNN features are also used like holistic features, but are extracted from sub-images which are the crops of raw images. In the CNN-based semantic segmentation paper \cite{hariharan2014simultaneous}, the R-CNN idea is generalized to masked raw image regions. For all these methods, the entire network is treated as a \emph{holistic} feature extractor, either on the entire image or on sub-images.

In the recent work of SPP-Net \cite{he2014spatial}, it shows that the convolutional feature maps can be used as \emph{localized} features. On a full-image convolutional feature map, the local rectangular regions encode both the semantic information (by strengths of activations) and spatial information (by positions). The features from these local regions can be pooled \cite{he2014spatial} directly for recognition.

The spatial pyramid pooling (SPP) in \cite{he2014spatial} actually plays two roles: 1) masking the feature maps by a rectangular region, outside which the activations are removed; 2) generating a fixed-length feature from this arbitrary sized region. So, if masking by rectangles can be effective, \emph{what if we mask the feature maps by a fine segment with an irregular shape}?

The Convolutional Feature Masking (CFM) layer is thus developed. We first obtain the candidate segments (like super-pixels) on the raw image. Many regional proposal methods (\eg, \cite{uijlings2013selective,arbelaez2014multiscale}) are based on super-pixels. Each proposal box is given by grouping a few super-pixels. We call such a group as a segment proposal. So we can obtain the candidate segments together with their proposal boxes (referred to as ``regions'' in this paper) without extra effort. These segments are binary masks on the raw images.

\begin{figure}
  \centering
     \includegraphics[width=0.8\linewidth]{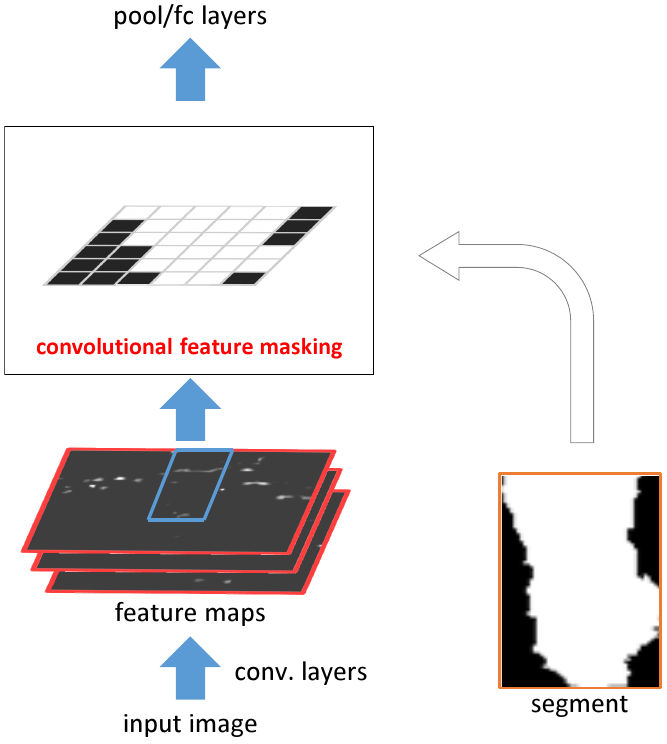}\\
  \caption{An illustration of the CFM layer.}
  \label{fig:feature_map_masking}
\end{figure}

Next we project these binary masks to the domain of the last convolutional feature maps. Because each activation in the convolutional feature maps is contributed by a receptive field in the image domain, we first project each activation onto the image domain as the center of its receptive field (following the details in \cite{he2014spatial}). Each pixel in the binary masks on the image is assigned to its nearest center of the receptive fields. Then these pixels are projected back onto the convolutional feature map domain based on this center and its activation's position. On the feature map, each position will collect multiple pixels projected from a binary mask. These binary values are then averaged and thresholded (by 0.5). This gives us a mask on the feature maps (Figure~\ref{fig:feature_map_masking}).
This mask is then applied on the convolutional feature maps. Actually, we only need to multiply this binary mask on each channel of the feature maps. We call the resulting features as \emph{segment features} in our method.

\subsection{Network Designs}

\begin{figure*}
  \centering
     \includegraphics[width=0.75\linewidth]{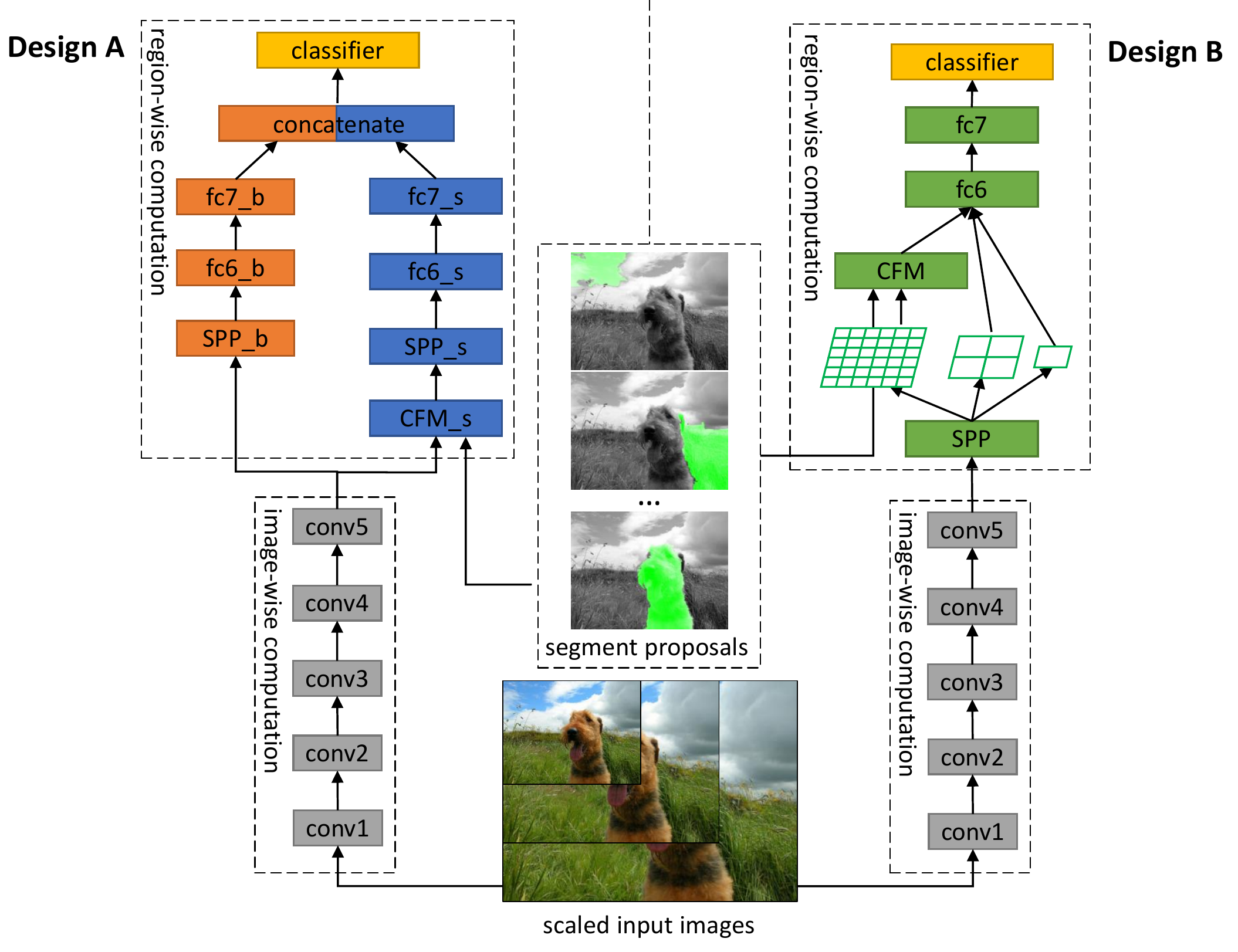}\\
  \caption{Two network designs in this paper. The input image is processed as a whole at the convolutional layers from conv1 to conv5. Segments are exploited at a deeper hierarchy by: (Left) applying CFM on the feature map of conv5, where ``\_b'' means for ``bounding boxes'' and ``\_s'' means for segments; (Right) applying CFM on the finest feature map of the spatial pyramid pooling layer.}
  \label{fig:network_architecture}
\end{figure*}

In \cite{hariharan2014simultaneous}, it has been shown that the segment features alone are insufficient. These segment features should be used together with the regional features (from bounding boxes) generated in a way like R-CNN \cite{girshick2013rich}. Based on our CFM layer, we can have two possible ways of doing this.

\vspace{6pt}
\noindent\textbf{Design A: on the last convolutional layer.} As shown in Figure \ref{fig:network_architecture} (left part), after the last convolutional layer, we generate two sources of features. One is the regional feature produced by the SPP layer as in \cite{he2014spatial}. The other is the segment feature produced in the following way. The CFM layer is applied on the full-image convolutional feature map. This gives us an arbitrary-sized (in terms of its bounding box) segment feature. Then we use another SPP layer on this feature to produce a fixed-length output. The two pooled features are fed into two separate fc layers. The features of the last fc layers are concatenated to train a classifier, as is the classifier in \cite{hariharan2014simultaneous}.

In this design, we have two pathways of the fc layers in both training and testing.

\vspace{6pt}
\noindent\textbf{Design B: on the spatial pyramid pooling layer.}
We first adopt the SPP layer \cite{he2014spatial} to pool the features. We use a 4-level pyramid of $\{6 \times 6, 3 \times 3, 2 \times 2, 1 \times 1\}$ as in \cite{he2014spatial}. The $6 \times 6$ level is actually a $6 \times 6$ tiny feature map that still has plenty spatial information. We apply the CFM layer on this tiny feature map to produce the segment feature. This feature is then concatenated with the other three levels and fed onto the fc layers, as shown in Figure \ref{fig:network_architecture} (right).

In this design, we keep one pathway of the fc layers to reduce the computational cost and over-fitting risk.

\subsection{Training and Inference}
\label{sec:training_and_inference}

Based on these two designs and the CFM layer, the training and inference stages can be easily conducted following the common practices in \cite{girshick2013rich,he2014spatial,hariharan2014simultaneous}. In both stages, we use the region proposal algorithm (\eg, selective search \cite{uijlings2013selective}) to generate about 2,000 region proposals and associated segments. The input image is resized to multiple scales (the shorter edge $s \in \{480, 576, 688, 864, 1200\}$) \cite{he2014spatial}, and the convolutional feature maps are extracted from full images and then fixed (not further tuned).

\vspace{6pt}
\noindent \textbf{Training.} We first apply the SPP method \cite{he2014spatial}\footnote{\url{https://github.com/ShaoqingRen/SPP_net}} to finetune a network for object detection. Then we replace the finetuned network with the architecture as in Design A or B, and further finetune the network for segmentation. In the second fine-tuning step, the segment proposal overlapping a ground-truth foreground segment by $[0.5,1]$ is considered as positive, and $[0.1,0.3]$ as negative. The overlap is measured by intersection-over-union (IoU) score based on the two segments' areas (rather than their bounding boxes).
After fine-tuning, we train a linear SVM classifier on the network output, for each category. In the SVM training, only the ground-truth segments are used as positive samples.

\vspace{6pt}
\noindent \textbf{Inference.}  Each region proposal is assigned to a proper scale as in \cite{he2014spatial}. The features of each region and its associated segment are extracted as in Design A or B. The SVM classifier is used to score each region.

Given all the scored region proposals, we obtain the pixel-level category labeling by the pasting scheme in SDS \cite{hariharan2014simultaneous}. This pasting scheme sequentially selects the region proposal with the highest score, performs region refinement, inhibits overlapping proposals, and pastes the pixel labels onto the labeling result. Region refinement improves the accuracy by about 1\% on PASCAL VOC 2012 for both SDS and our method.

\subsection{Results on Object Segmentation}
\label{sec:exp1}

We evaluate our method on the PASCAL VOC 2012 semantic segmentation benchmark \cite{everingham2010pascal} that has 20 object categories. We follow the ``comp6" evaluation protocol, which is also used in \cite{carreira2012semantic,girshick2013rich,hariharan2014simultaneous}. The training set of PASCAL VOC 2012 and the additional segmentation annotations from \cite{hariharan2011semantic} are used for training and evaluation as in \cite{carreira2012semantic,girshick2013rich,hariharan2014simultaneous}. Two scenarios are studied: semantic segmentation and simultaneous detection and segmentation.

\subsubsection*{Scenario I: Semantic Segmentation}

In the experiments of semantic segmentation, category labels are assigned to all the pixels in the image, and the accuracy is measured by region IoU scores \cite{everingham2010pascal}.

We first study using the ``ZF SPPnet'' model \cite{he2014spatial} as our feature extractor. This model is based on Zeiler and Fergus's fast model \cite{zeiler2013visualizing} but with the SPP layer \cite{he2014spatial}. It has five convolutional layers and three fc layers. This model is released with the code of \cite{he2014spatial}. We note that the results in R-CNN \cite{girshick2013rich} and SDS \cite{hariharan2014simultaneous} use the ``AlexNet'' \cite{krizhevsky2012imagenet} instead. To understand the impacts of the pre-trained models, we report their \emph{object detection} mAP on the \emph{val} set of PASCAL VOC 2012: SPP-Net (ZF) is 51.3\%, R-CNN (AlexNet) is 51.0\%, and SDS (AlexNet) is 51.9\%. This means that both pre-trained models are comparable as generic feature extractors. So the following gains of CFM are not simply due to pre-trained models.

\begin{table}[t]
\renewcommand{\arraystretch}{1.2}
\begin{center}
\small
\begin{tabular}{x|x|x}
\hline
              no-CFM~   &CFM (A)  &CFM (B)   \\
\hline
          43.4          & 51.0       & 50.9      \\
\hline
\end{tabular}
\end{center}
\caption{Mean IoU on PASCAL VOC 2012 {\em validation} set using our various designs. Here we use ZF SPPnet and Selective Search.}
\label{tab:voc2012_network_architecture1}
\vspace{6pt}
\renewcommand{\arraystretch}{1.2}
\begin{center}
\small
\begin{tabular}{x|x|x}
\hline
              ~   & ZF SPPnet & VGG net \\
\hline
          SS      & 50.9 & 56.3      \\
          MCG     & 53.0 & 60.9      \\
\hline
\end{tabular}
\end{center}
\caption{Mean IoU on PASCAL VOC 2012 {\em validation} set using different pre-trained networks and proposal methods. SS denotes Selective Search \cite{uijlings2013selective}, and MCG denotes Multiscale Combinatorial Grouping \cite{arbelaez2014multiscale}.}
\label{tab:voc2012_network_architecture2}
\vspace{6pt}
\renewcommand{\arraystretch}{1.2}
\begin{center}
\small
\begin{tabular}{x|x|x}
\hline
              ~   & ZF SPPnet  & VGG net   \\
\hline
          5-scale      & 53.0 & 60.9      \\
          1-scale      & 52.9 & 60.5      \\
\hline
\end{tabular}
\end{center}
\caption{Mean IoU on PASCAL VOC 2012 {\em validation} set using different scales. Here we use MCG for proposals.}
\label{tab:voc2012_network_architecture3}
\vspace{6pt}
\begin{center}
\renewcommand{\arraystretch}{1.2}
\small
\begin{tabular}{c|cc|c}
\hline
                  &conv time     &fc time  &total time        \\
\hline
SDS (AlexNet) \cite{hariharan2014simultaneous}&17.8s              &0.14s              &17.9s      \\
\hline
CFM, (ZF, 5 scales)  &{0.29s}     &{0.09s}     &{0.38s}         \\
CFM, (ZF, 1 scale)  &{0.04s}     &{0.09s}     &{0.12s}         \\
CFM, (VGG, 5 scales)  & 1.74s     & 0.36s     & 2.10s      \\
CFM, (VGG, 1 scale)  & 0.21s     & 0.36s     & 0.57s      \\
\hline
\end{tabular}
\end{center}
\caption{Feature extraction time per image on GPU.}
\label{fig:voc2012_speed}
\end{table}

To show the effect of the CFM layer, we present a baseline with no CFM - in our Design B, we remove the CFM layer but still use the same entire pipeline. We term this baseline as the ``no-CFM'' version of our method.
Actually, this baseline degrades to the original SPP-net usage \cite{he2014spatial}, except that
the definitions of positive/negative samples are for segmentation. Table \ref{tab:voc2012_network_architecture1} compares the results of no-CFM and the two designs of CFM. We find that the CFM has obvious advantages over the no-CFM baseline. This is as expected, because the no-CFM baseline has not any segment-based feature. Further, we find that the designs A and B perform just comparably, while A needs to compute two pathways of the fc layers. So in the rest of this paper, we adopt Design B for ZF SPPnet.

In Table \ref{tab:voc2012_network_architecture2} we evaluate our method using different region proposal algorithms. We adopt two proposal algorithms: Selective Search (SS) \cite{uijlings2013selective}, and Multiscale Combinatorial Grouping (MCG) \cite{arbelaez2014multiscale}.
Following the protocol in \cite{hariharan2014simultaneous}, the ``fast" mode is used for SS, and the ``accurate'' mode is used for MCG. Table \ref{tab:voc2012_network_architecture2} shows that our method achieves higher accuracy on the MCG proposals. This indicates that our feature masking method can exploit the information generated by more accurate segmentation proposals.

In Table \ref{tab:voc2012_network_architecture2} we also evaluate the impact of pre-trained networks.
We compare the ZF SPPnet with the public VGG-16 model \cite{simonyan2014very}\footnote{\url{www.robots.ox.ac.uk/~vgg/research/very_deep/}}.
Recent advances in image classification have shown that very deep networks \cite{simonyan2014very} can significantly improve the classification accuracy. The VGG-16 model has 13 convolutional and 3 fc layers. Because this model has no SPP layer, we consider its last pooling layer ($7\times7$) as a special SPP layer which has a single-level pyramid of $\{7 \times 7\}$. In this case, our Design B does not apply because there is no coarser level. So we apply our Design A instead. Table \ref{tab:voc2012_network_architecture2} shows that our results improve substantially when using the VGG net. This indicates that our method benefits from the more representative features learned by deeper models.

In Table \ref{tab:voc2012_network_architecture3} we evaluate the impact of image scales. Instead of using the 5 scales, we simply extract features from single-scale images whose shorter side is $s=576$. Table \ref{tab:voc2012_network_architecture3} shows that our single-scale variant has negligible degradation. But the single-scale variant has a faster computational speed as in Table~\ref{fig:voc2012_speed}.

\begin{table*}[t]
\begin{center}
\setlength{\tabcolsep}{2.1pt}
\renewcommand{\arraystretch}{1.2}
\resizebox{\linewidth}{!}{
\begin{tabular}{x|x|xxxxxxxxxxxxxxxxxxxx}
  \hline
  & mean & \ct{areo} & \ct{bike} & \ct{bird} & \ct{boat} & \ct{bottle} & \ct{bus} & \ct{car} & \ct{cat} & \ct{chair} & \ct{cow} & \ct{table} & \ct{dog} & \ct{horse} & \ct{mbike} & \ct{person} & \ct{plant} & \ct{sheep} & \ct{sofa} & \ct{train} & \ct{tv}\\
  \hline
  O$_2$P \cite{carreira2012semantic}&47.8&64.0&\textbf{27.3}&54.1&39.2&48.7&56.6&57.7&52.5&14.2&54.8&29.6&42.2&58.0&54.8&50.2&36.6&58.6&31.6&48.4&38.6 \\
  SDS (AlexNet + MCG) \cite{hariharan2014simultaneous}&51.6&63.3&25.7&63.0&39.8&59.2&{70.9}&61.4&54.9&16.8&45.0&48.2&50.5&51.0&57.7&{63.3}&31.8&58.7&31.2&{55.7}&48.5
  \\
  \hline
  CFM (ZF + SS)&53.5&63.3&21.5&59.1&40.3&52.4&68.6&55.4&66.6&25.4&60.5&48.5&60.0&53.6&58.6&59.8&40.5&\textbf{68.6}&31.7&49.3&53.6\\
  CFM (ZF + MCG)&55.4&65.2&23.5&59.0&{40.4}&61.1&68.9&57.9&70.8&23.9&59.4&44.7&66.2&57.5&62.1&57.6&44.1&{64.5}&42.5&52.9&\textbf{55.7}\\
  CFM (VGG + MCG)&{\textbf{61.8}}&{\textbf{75.7}}&{26.7}&{\textbf{69.5}}&\textbf{48.8}&{\textbf{65.6}}&\textbf{81.0}&{\textbf{69.2}}&{\textbf{73.3}}&{\textbf{30.0}}&{\textbf{68.7}}&{\textbf{51.5}}&{\textbf{69.1}}&{\textbf{68.1}}&{\textbf{71.7}}&\textbf{67.5}&{\textbf{50.4}}&{66.5}&{\textbf{44.4}}&\textbf{58.9}&{53.5}\\
  \hline
\end{tabular}
}
\end{center}
\caption{Mean IoU scores on the PASCAL VOC 2012 {\em test} set.}
\label{tab:voc2012_approaches}
\end{table*}

Next we compare with the state-of-the-art results on the PASCAL VOC 2012 test set in Table \ref{tab:voc2012_approaches}. Here SDS \cite{hariharan2014simultaneous} is the previous state-of-the-art method on this task, and O$_2$P \cite{carreira2012semantic} is a leading non-CNN-based method. Our method with ZF SPPnet and MCG achieves a score of 55.4. This is 3.8\% higher than the SDS result reported in \cite{hariharan2014simultaneous} which uses AlexNet and MCG. This demonstrates that our CFM method can produce effective features without masking raw-pixel images. With the VGG net, our method has a score of \textbf{61.8} on the test set.

Besides the high accuracy, our method is much faster than SDS. The running time of the feature extraction steps in SDS and our method is shown in Table~\ref{fig:voc2012_speed}. Both approaches are run on an Nvidia GTX Titan GPU based on the Caffe library \cite{jia2014caffe}. The time is averaged over 100 random images from PASCAL VOC. Using 5 scales, our method with ZF SPPnet is $\sim47\times$ faster than SDS; using 1 scale, our method with ZF SPPnet is $\sim$$150\times$ faster than SDS and is more accurate. The speed gain is because our method only needs to compute the feature maps once. Table~\ref{fig:voc2012_speed} also shows that our method is still feasible using the VGG net.

Concurrent with our work, a Fully Convolutional Network (FCN) method \cite{Long2014} is proposed for semantic segmentation. It has a score (62.2 on test set) comparable with our method, and has a fast speed as it also performs convolutions once on the entire image. But FCN is not able to generate instance-wise results, which is another metric evaluated in \cite{hariharan2014simultaneous}. Our method is also applicable in this case, as evaluated below.

\subsubsection*{Scenario II: Simultaneous Detection and Segmentation}

In the evaluation protocol of simultaneous detection and segmentation \cite{hariharan2014simultaneous}, all the object instances and their segmentation masks are labeled. In contrast to semantic segmentation, this scenario further requires to identify different object instances in addition to labeling pixel-wise semantic categories. The accuracy is measured by mean AP\textsuperscript{r} score defined in \cite{hariharan2014simultaneous}.

We report the mean AP\textsuperscript{r} results on VOC 2012 validation set following \cite{hariharan2014simultaneous}, as the ground-truth labels for the test set are not available.
As shown in Table \ref{tab:voc2012_approaches_sds}, our method has a mean AP\textsuperscript{r} of 53.2 when using ZF SPPnet and MCG. This is better than the SDS result (49.7) reported in \cite{hariharan2014simultaneous}. With the VGG net, our mean AP\textsuperscript{r} is \textbf{60.7}, which is the state-of-the-art result reported in this task. Note that the FCN method \cite{Long2014} is not applicable when evaluating the mean AP\textsuperscript{r} metric, because it cannot produce object instances.

\begin{table}[t]
\begin{center}
\renewcommand{\arraystretch}{1.2}
\small
\begin{tabular}{c|c}
\hline
method & mean AP\textsuperscript{r}\\
\hline
SDS (AlexNet + MCG) \cite{hariharan2014simultaneous} & 49.7\\
\hline
CFM (ZF + SS) & 51.0\\
CFM (ZF + MCG) & 53.2\\
CFM (VGG + MCG) & \textbf{60.7}\\
\hline
\end{tabular}
\end{center}
\caption{Instance-wise semantic segmentation evaluated by mean AP\textsuperscript{r} \cite{hariharan2014simultaneous} on PASCAL VOC 2012 {\em validation} set.}
\label{tab:voc2012_approaches_sds}
\end{table}

\section{Joint Object and Stuff Segmentation}

The semantic categories in natural images can be roughly divided into \emph{objects} and \emph{stuff}. Objects have consistent shapes and each instance is countable, while stuff has consistent colors or textures and exhibits as arbitrary shapes, \eg, grass, sky, and water. So unlike an object, a stuff region is not appropriate to be represented as a rectangular region or a bounding box.
While our method can generate segment features, each segment is still associated with a bounding box due to its way of generation. When the region/segment proposals are provided, it is rarely that the stuff can be fully covered by a single segment. Even if the stuff is covered by a single rectangular region, it is almost certain that there are many pixels in this region that do not belong to the stuff. So stuff segmentation has issues different from object segmentation.

Next we show a generalization of our framework to address this issue involving stuff. We can simultaneously handle objects and stuff by a single solution. Especially, the convolutional feature maps need only to be computed once. So there will be little extra cost if the algorithm is required to further handle stuff.

Our generalization is to modify the underlying probabilistic distributions of the samples during training. Instead of treating the samples equally, our training will bias toward the proposals that can cover the stuff as compact as possible (discussed below). A Segment Pursuit procedure is proposed to find the compact proposals.

\subsection{Stuff Representation by Segment Combination}

We treat stuff as a combination of multiple segment proposals. We expect that each segment proposal can cover a stuff portion as much as possible, and a stuff can be fully covered by several segment proposals. At the same time, we hope the combination of these segment proposals is compact - the fewer the segments, the better.

We first define a candidate set of segment proposals (in a single image) for stuff segmentation.
We define a ``purity score'' as the IoU ratio between a segment proposal and the stuff portion that is within the bounding box of this segment. Among all the segment proposals in a single image, those having high purity scores ($>0.6$) with stuff consist of the candidate set for potential combinations.

To generate one compact combination from this candidate set, we adopt a procedure similar to the matching pursuit \cite{wu2010learning,mallat1993matching}. We sequentially pick segments from the candidate set without replacement. At each step, the largest segment proposal is selected. This selected proposal then inhibits its highly overlapped proposals in the candidate set (they will not be selected afterward). In this paper, the inhibition overlap threshold is set as IoU=0.2.
The process is repeated till the remaining segments all have areas smaller than a threshold, which is the average of the segment areas in the initial candidate set (of that image). We call this procedure segment pursuit.

\begin{figure}[t]
  \centering
    \includegraphics[width=0.9\linewidth]{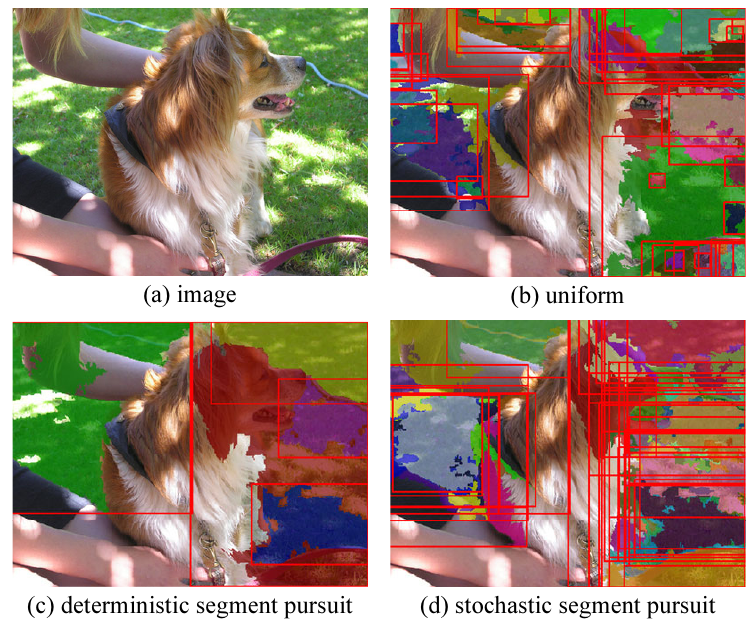}
  \caption{Stuff segment proposals sampled by different methods. (a) input image; (b) 43 regions uniformly sampled; (c) 5 regions sampled by deterministic segment pursuit; (d) 43 regions sampled by stochastic segment pursuit for finetuning.}
  \label{fig:biased_sampling}
\end{figure}

Figure \ref{fig:biased_sampling} (b) shows an example if segment proposals are randomly sampled from the candidate set. We see that there are many small segments. It is harmful to define these small, less discriminative segments as either positive or negative samples (\eg, by IoU) - if they are positive, they are just a very small part of the stuff; if they are negative, they share the same textures/colors as a larger portion of the stuff. So we prefer to ignore these samples in the training, so the classifier will not bias toward any side about these small samples.
Figure \ref{fig:biased_sampling} (c) shows the segment proposals selected by segment pursuit. We see that they can cover the stuff (grass here) by only a few but large segments. We expect the solver to rely more on such a compact combination of proposals.

However, the above process is \emph{deterministic} and can only give a small set of samples from each image. For example, in Figure \ref{fig:biased_sampling} (c) it only provides 5 segment proposals.
In the fine-tuning process, we need a large number of \emph{stochastic} samples for the training. So we inject randomness into the above segment pursuit procedure. In each step, we randomly sample a segment proposal from the candidate set, rather than using the largest. The picking probability is proportional to the area size of a segment (so a larger one is still preferred). This can give us another compact combination in a stochastic way. Figure \ref{fig:biased_sampling} (d) shows an example of the segment proposals generated in a few trials.

All the segment proposals given by this way are considered as the positive samples of a category of stuff. The negative samples are the segment proposals whose purity scores are below 0.3. These samples can then be used for fine-tuning and SVM training as detailed below.

\begin{table}
\setlength{\tabcolsep}{6pt}
\newcommand{\hl}[1]{\Large{#1}}
\newcommand{\hll}[1]{\large{#1}}
\newcommand{\tabincell}[2]{\begin{tabular}{@{}#1@{}}#2\end{tabular}}
\renewcommand{\arraystretch}{1.05}
\begin{center}
\resizebox{1.\linewidth}{!}{
\begin{tabular}{|x|xx|xxx|x|x|}
	\hline
	& & & & & & &\\[-2ex]
	& & &\multicolumn{3}{c|}{\hll{ZF+SS}} &\hll{VGG+SS} &\hll{VGG+MCG}\\
	\hline
	& & & & & & & \\[-1.5ex]
	&\hll{\tabincell{c}{Super\\Parsing}}   &\hll{O$_2$P} & \hll{no-CFM} & \hll{\tabincell{c}{CFM \\w/o SP}} &\hll{CFM} &\hll{CFM} &\hll{CFM}\\
	& & & & & & & \\[-2ex]
	\hline
	& & & & & & & \\[-2ex]
	\hl{mean}            &   - & \hl{18.1} & \hl{20.7} & \hl{24.0} & \hl{26.6} &\hl{{31.5}}&\textbf{\hl{34.4}}\\[4pt]
	\hl{mean on \dag}    &   \hl{15.2} & \hl{29.2} & \hl{32.4} & \hl{37.2} & \hl{40.4} & \hl{46.1}&\textbf{\hl{49.5}}\\
	& & & & & & & \\[-2.5ex]
	\hline
	aeroplane\dag&19.5&36.4&20.5&37.6&42.9&{\textbf{48.9}}&47.5\\
	bicycle\dag&11.3&23.5&32.2&39.1&{40.3}&{41.2}&\textbf{48.0}\\
	bird\dag&4.1&24.6&27.9&40.5&{46.6}&{52.9}&\textbf{59.0}\\
	boat\dag&0.0&22.3&16.6&29.9&34.0&{33.6}&\textbf{37.7}\\
	bottle\dag&1.2&15.0&{40.0}&39.0&39.8&41.5&\textbf{51.6}\\
	bus\dag&14.0&43.2&50.0&52.4&{53.5}&{61.0}&\textbf{65.2}\\
	car\dag&15.0&33.5&41.0&44.0&{47.1}&53.7&\textbf{57.2}\\
	cat\dag&20.1&36.7&45.1&54.1&{56.1}&60.0&\textbf{67.4}\\
	chair\dag&2.9&6.8&13.6&15.7&{17.7}&22.9&\textbf{24.6}\\
	cow\dag&0.1&16.2&28.5&34.8&{39.8}&{52.4}&\textbf{58.9}\\
	table\dag&6.4&7.0&12.1&12.3&13.9&{11.5}&\textbf{16.7}\\
	dog\dag&11.5&26.9&39.5&48.3&{51.4}&57.6&\textbf{63.7}\\
	horse\dag&2.0&26.4&33.0&40.2&{43.1}&{\textbf{50.5}}&\textbf{58.0}\\
	motorbike\dag&14.3&32.8&40.4&45.1&{47.9}&54.8&\textbf{55.0}\\
	person\dag&30.1&44.5&47.4&51.0&{54.5}&59.9&\textbf{65.0}\\
	pottedplant\dag&1.1&15.9&31.4&31.5&34.9&{34.1}&\textbf{41.1}\\
	sheep\dag&4.2&23.7&29.3&45.5&{56.3}&59.6&\textbf{60.7}\\
	sofa\dag&3.6&16.1&15.2&19.1&{22.0}&22.1&\textbf{31.8}\\
	train\dag&10.4&26.7&33.6&39.1&{43.0}&{49.0}&\textbf{56.1}\\
	tvmonitor\dag&9.0&24.3&40.3&{41.0}&40.7&{\textbf{50.4}}&50.3\\
	sky\dag&65.6&75.6&64.9&70.3&{76.8}&{\textbf{80.6}}&76.8\\
	grass\dag&45.3&56.0&51.9&56.9&{60.7}&{\textbf{66.1}}&66.1\\
	ground\dag&24.0&{27.6}&22.0&20.5&22.8&38.3&\textbf{39.4}\\
	road\dag&15.8&31.2&25.3&30.6&{34.0}&{36.2}&\textbf{37.8}\\
	building\dag&19.8&24.3&25.0&28.2&{32.4}&37.9&\textbf{39.5}\\
	tree\dag&37.8&44.3&44.2&50.9&{53.4}&{\textbf{59.8}}&58.0\\
	water\dag&34.5&54.8&51.4&54.1&{59.7}&65.3&\textbf{69.1}\\
	mountain\dag&8.8&19.2&14.9&{20.8}&18.4&{\textbf{36.7}}&35.6\\
	wall\dag&30.8&{40.5}&28.7&36.4&40.4&{42.5}&\textbf{43.8}\\
	floor\dag&14.4&25.7&23.0&28.1&{31.7}&{35.9}&\textbf{38.9}\\
	track\dag&17.5&29.5&{35.7}&27.3&31.9&{\textbf{40.1}}&38.2\\
	keyboard\dag&0.1&18.2&26.1&{30.2}&25.1&{36.7}&\textbf{39.8}\\
	ceiling\dag&6.4&12.7&19.3&13.4&{20.3}&\textbf{27.9}&23.8\\
	bag&-&1.2&0.5&2.1&2.8&{2.1}&\textbf{9.0}\\
	bed&-&0.7&0.0&1.2&\textbf{3.0}&{1.1}&2.9\\
	bedclothes&-&0.0&4.0&9.9&{11.9}&13.7&\textbf{16.6}\\
	bench&-&0.1&0.0&0.0&0.1&{0.0}&\textbf{0.2}\\
	book&-&5.0&{15.0}&8.9&14.8&{\textbf{20.5}}&20.1\\
	cabinet&-&4.4&4.2&5.0&{8.7}&11.4&\textbf{18.4}\\
	cloth&-&1.8&0.2&2.8&\textbf{3.4}&{2.7}&2.7\\
	computer&-&0.0&0.0&3.8&5.0&{4.5}&\textbf{8.6}\\
	cup&-&1.4&7.4&8.1&{12.4}&21.2&\textbf{25.5}\\
	curtain&-&11.6&12.5&9.2&{15.5}&21.4&\textbf{25.1}\\
	door&-&2.3&3.9&5.0&{6.2}&{\textbf{12.7}}&4.9\\
	fence&-&6.6&{8.4}&4.9&7.3&20.7&\textbf{23.3}\\
	flower&-&6.8&4.1&{9.0}&8.4&{10.7}&\textbf{28.0}\\
	food&-&10.7&22.7&23.3&{29.8}&35.2&\textbf{38.1}\\
	mouse&-&0.9&1.4&4.4&{9.9}&\textbf{12.4}&12.3\\
	plate&-&5.6&7.5&7.0&{10.7}&19.3&\textbf{21.8}\\
	platform&-&7.5&14.7&16.2&{18.4}&18.7&\textbf{26.7}\\
	rock&-&6.7&8.1&13.5&{15.0}&24.2&\textbf{26.4}\\
	shelves&-&{3.7}&1.8&1.5&3.6&{3.1}&\textbf{10.4}\\
	sidewalk&-&0.5&0.0&1.7&{2.5}&{\textbf{7.9}}&6.8\\
	sign&-&7.0&1.6&4.1&{8.7}&\textbf{18.7}&17.2\\
	snow&-&16.4&19.0&23.9&{28.9}&{28.3}&\textbf{40.5}\\
	truck&-&0.2&0.3&0.4&{4.5}&{3.4}&\textbf{11.0}\\
	window&-&{14.6}&10.7&11.9&12.4&{\textbf{21.7}}&19.2\\
	wood&-&0.8&0.5&2.4&2.8&{1.9}&\textbf{5.8}\\
	light&-&8.5&{11.5}&6.2&10.9&{15.3}&\textbf{20.5}\\
	\hline
\end{tabular}}
\end{center}
\caption{Segmentation accuracy measured by IoU scores on the new PASCAL-CONTEXT validation set \cite{mottaghi2014role}. The categories marked by \dag~are the 33 easier categories identified in \cite{mottaghi2014role}. The results of SuperParsing \cite{tighe2010superparsing} and O$_2$P \cite{carreira2012semantic} are from the \emph{errata} of \cite{mottaghi2014role}.}
\label{tab:voc2010_approaches}
\end{table}

\begin{figure*}[t]
  \centering
  \includegraphics[width=0.9\textwidth]{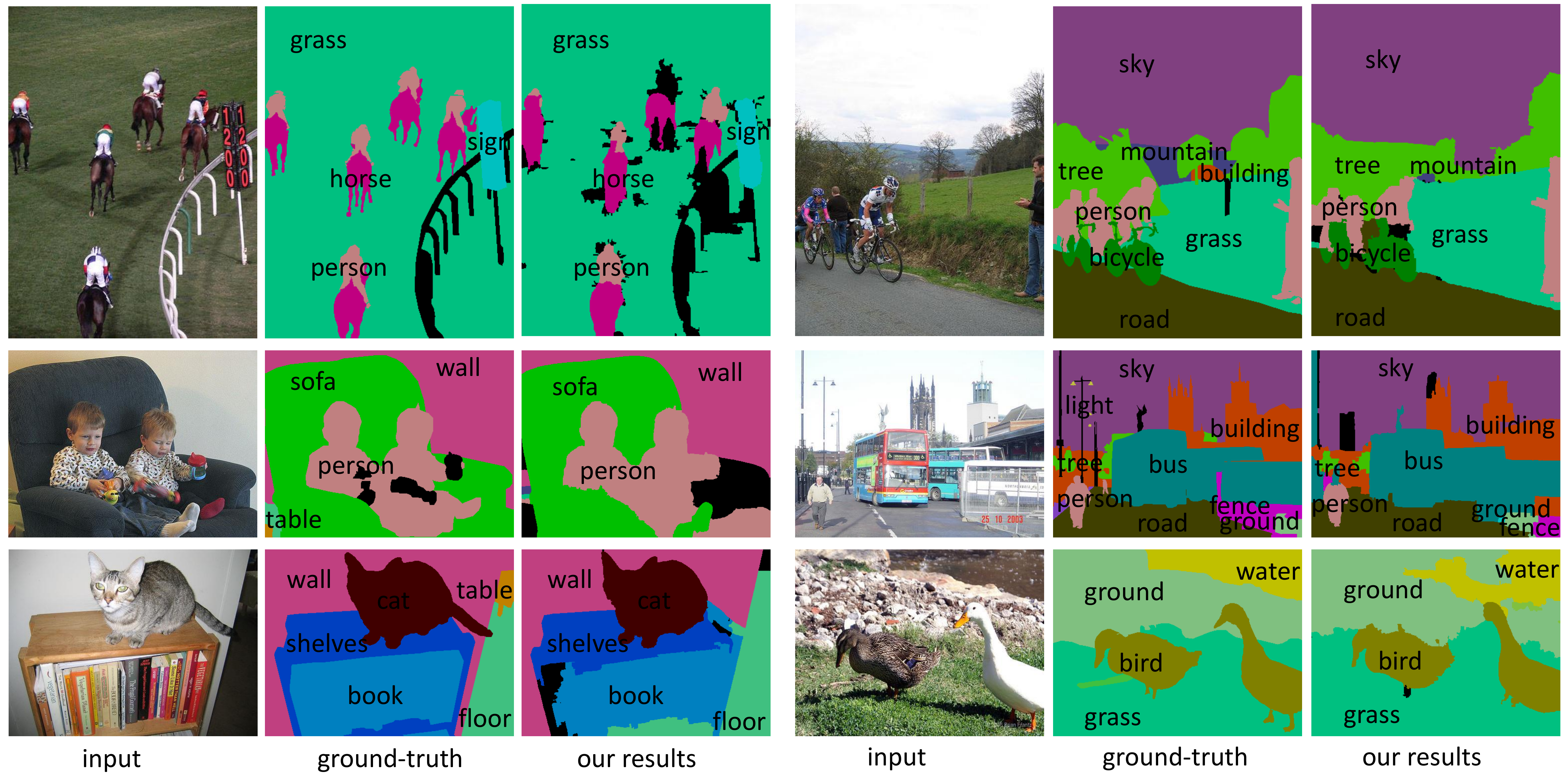}
  \caption{Some example results of our CFM method (with VGG and MCG) for \textbf{joint object and stuff segmentation}. The images are from the PASCAL-CONTEXT validation set \cite{mottaghi2014role}.}
  \label{fig:segmentation_examples}
\end{figure*}

During the fine-tuning stage, in each epoch each image generates a stochastic compact combination. All the segment proposals in this combination for all images consist of the samples of this epoch. These samples are randomly permuted and fed into the SGD solver. Although now the samples appear mutually independent to the SGD solver, they are actually sampled jointly by the rule of segment pursuit. Their underlying probabilistic distributions will impact the SGD solver. This process is repeated for each epoch. For the SGD solver, we halt the training process after 200k mini-batches. For SVM training, we only use the single combination given by the deterministic segment pursuit.

Using this way, we can treat object+stuff segmentation in the same framework as for object-only. The only difference is that the stuff samples are provided in a way given by segment pursuit, rather than purely randomly.
To balance different categories, the portions of objects, stuff, and background samples in each mini-batch are set to be approximately 30\%, 30\%, and 40\%. The testing stage is the same as in the object-only case. While the testing stage is unchanged, the classifiers learned are biased toward those compact proposals.

\subsection{Results on Joint Object and Stuff Segmentation}

We conduct experiments on the newly labeled PASCAL-CONTEXT dataset \cite{mottaghi2014role} for joint object and stuff segmentation. In this enriched dataset, every pixel is labeled with a semantic category. It is a challenging dataset with various images, diverse semantic categories, and balanced ratios of object/stuff pixels. Following the protocol in \cite{mottaghi2014role}, the semantic segmentation is performed on the most frequent 59 categories and one background category (Table \ref{tab:voc2010_approaches}). The segmentation accuracy is measured by mean IoU scores over the 60 categories. Following \cite{mottaghi2014role}, the mean of the scores over a subset of 33 easier categories (identified by \cite{mottaghi2014role}) is reported in this 60-way segmentation task as well. The training and evaluation are performed on the train and val sets respectively.
We compare with two leading methods - SuperParsing \cite{tighe2010superparsing} and O$_2$P \cite{carreira2012semantic}, whose results are reported in \cite{mottaghi2014role}. For fair comparisons, the region refinement \cite{hariharan2014simultaneous} is not used in all methods. The pasting scheme is the same as in O$_2$P \cite{carreira2012semantic}. In this comparison, we ignore R-CNN \cite{girshick2013rich} and SDS \cite{hariharan2014simultaneous} because they have not been developed for stuff.

Table \ref{tab:voc2010_approaches} shows the mean IoU scores. Here ``no-CFM'' is our baseline (no CFM, no segment pursuit); ``CFM w/o SP'' is our CFM method but without segment pursuit; and ``CFM'' is our CFM method with segment pursuit. When segment pursuit is not used, the positive stuff samples are uniformly sampled from the candidate set (in which the segments have purity scores $>$ 0.6).

SuperParsing \cite{tighe2010superparsing} gets a mean score of 15.2 on the easier 33 categories, and the overall score is unavailable in \cite{mottaghi2014role}. The O$_2$P method \cite{carreira2012semantic} results in 29.2 on the easier 33 categories and 18.1 overall, as reported in \cite{mottaghi2014role}. Both methods are not based on CNN features.

For the CNN-based results, the no-CFM baseline (20.7, with ZF and SS) is already better than O$_2$P (18.1). This is mainly due to the generic features learned by deep networks. Our CFM method without segment pursuit improves the overall score to 24.0. This shows the effects of the masked convolutional features. With our segment pursuit, the CFM method further improves the overall score to 26.6. This justifies the impact of the samples generated by segment pursuit. When replacing the ZF SPPnet by the VGG net, and the SS proposals by MCG, our method yields an over score of \textbf{34.4}. So our method benefits from deeper models and more accurate segment proposals.
Some of our results are shown in Figure~\ref{fig:segmentation_examples}.

It is worth noticing that although only mean IoU scores are evaluated in this dataset, our method is also able to generate instance-wise results for objects.

\begin{figure*}[t]
\centering
\includegraphics[width=0.92\linewidth]{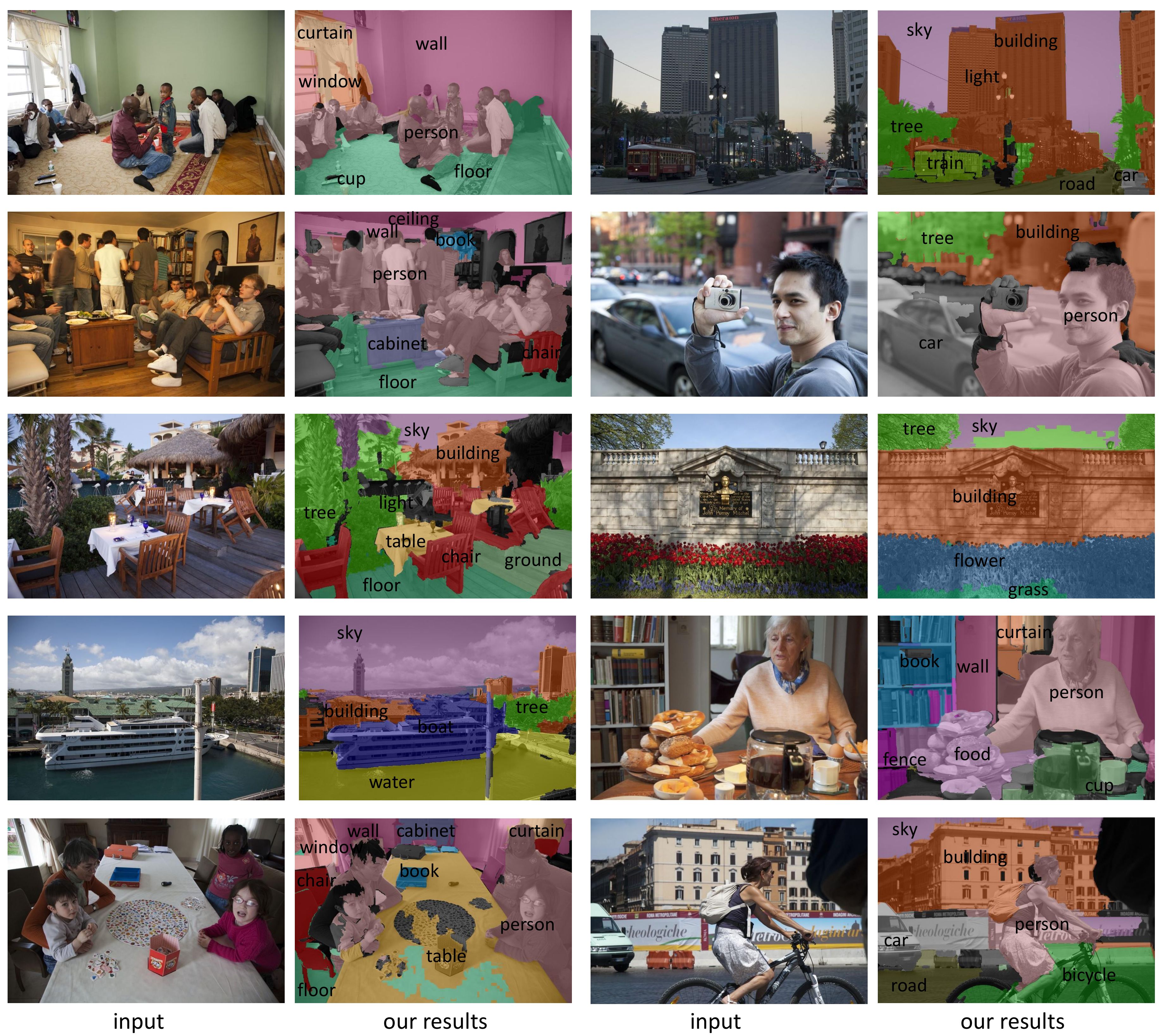}
\caption{Some visual results of our trained model (with VGG and MCG) for \textbf{cross-dataset joint object and stuff segmentation}. The network is trained on the PASCAL-CONTEXT training set \cite{mottaghi2014role}, and is applied on MIT-Adobe FiveK \cite{vladimir2011fivek}.}
\label{fig:results_mit5k}
\end{figure*}

\vspace{6pt}
\noindent\textbf{Additional Results.}
We also run our trained model on an \emph{external} dataset of MIT-Adobe FiveK \cite{vladimir2011fivek}, which consists of images taken by professional photographers to cover a broad range of scenes, subjects, and lighting conditions. Although our model is not trained for this dataset, it produces reasonably good results (see Figure \ref{fig:results_mit5k}).

\section{Conclusion}

We have presented convolutional feature masking, which exploits the shape information at a late stage in the network. We have further shown that convolutional feature masking is applicable for joint object and stuff segmentation.

We plan to further study improving object detection by convolutional feature masking.
Exploiting the context information provided by joint object and stuff segmentation would also be interesting.

{\small
\bibliographystyle{ieee}
\bibliography{seg_3}
}

\end{document}